\documentclass[runningheads]{llncs}

\usepackage{amssymb}
\usepackage{amsmath, graphicx}

\usepackage[dvipsnames]{xcolor}
\usepackage[ruled]{algorithm2e}
\usepackage{setspace}

\usepackage{empheq}
\usepackage{amssymb}
\usepackage{mathabx}
\usepackage{array}
\usepackage{float}
\usepackage{subfig}
\usepackage{pgf,tikz}
\usepackage{multicol}
\usepackage{multirow}
\usepackage{rotating}
\usepackage{booktabs}
\usepackage{graphicx}
\begin{document}
\title{Deep Convolutional Transform Learning\thanks{This work was supported by the CNRS-CEFIPRA project under grant NextGenBP PRC2017.}}
%
%
\author{Jyoti Maggu\inst{1} \and
Angshul Majumdar\inst{1}\and
Emilie Chouzenoux\inst{2}
\and
Giovanni Chierchia\inst{3}}
%
\authorrunning{J. Maggu et al.}
%
\institute{Indraprastha Institute of Information Technology Delhi, India.\\
\email{{jyotim, angshul}@iiitd.ac.in}\\
\and
CVN, Inria Saclay, Univ. Paris-Saclay, CentraleSup\'elec,  Gif-sur-Yvette, France.\\
\email{emilie.chouzenoux@centralesupelec.fr}
\and
LIGM, ESIEE Paris, Univ. Gustave Eiffel, Noisy-le-Grand, France.\\
\email{giovanni.chierchia@esiee.fr}}
\maketitle              
\begin{abstract}
This work introduces a new unsupervised representation learning technique called Deep Convolutional Transform Learning (DCTL). By stacking convolutional transforms, our approach is able to learn a set of independent kernels at different layers. The features extracted in an unsupervised manner can then be used to perform machine learning tasks, such as classification and clustering. The learning technique relies on a well-sounded alternating proximal minimization scheme with established convergence guarantees. Our experimental results show that the proposed DCTL technique outperforms its shallow version CTL, on several benchmark datasets.

\keywords{Transform Learning \and Deep Learning \and  Convolutional Neural Networks \and  Classification \and  Clustering \and  Proximal Methods \and  Alternating Minimization}
\end{abstract}
\section{Introduction}
Deep learning and more particularly convolutional neural networks (CNN) have penetrated almost every perceivable area of signal/image processing and machine learning. Its performance in traditional machine learning tasks encountered in computer vision, natural language processing and speech analysis are well assessed. CNNs are also being used with success in traditional signal processing domains, such as biomedical signal analysis \cite{hannun2019cardiologist}, radars \cite{mason2017deep}, astronomy \cite{chan2019deep} and inverse problems \cite{ye2018deep}. When large volumes of labeled data are available, CNNs can be trained efficiently using back-propagation methods and reach excellent performance~ \cite{rumelhart1986learning}. However, training a CNN requires labeled data in a large quantity. The latter issue can be overcome by considering alternate learning paradigms, such as spiking neural network (SNN) \cite{van2017artificial} and the associated Hebbian learning \cite{kempter1999hebbian}, or alternate optimization strategies such as in \cite{taylor2016training}. However, none of those approaches can overcome the fundamental problem of neural networks, that is their limited capacity of learning in an unsupervised fashion. This explains the great recent interest in the machine learning community for investigating representation learning methods, that keep the best of both worlds, that is the performance of multi-layer convolutional representations and the unsupervised learning capacity \cite{Chabiron2015,Tang2020,Fagot2019,maggu2018convolutional,ElGheche2019_tsipn}. 

In this work, we propose a deep version of the convolutional transform learning (CTL) approach introduced in \cite{maggu2018convolutional}, that we call deep convolutional transform learning (DCTL). A proximal alternating minimization scheme allows us to learn multiple layers of convolutional filters in an unsupervised fashion. Numerical experiments illustrate the ability of the method to learn representative features that lead to great performance on a set of classification and clustering problems. 


The rest of the paper is organized into several sections. Section \ref{literature} introduces the transform learning paradigm and briefly reminds our previous CTL approach. The proposed DCTL formulation and the associated learning strategy are presented in Section \ref{proposed}. The experimental results are described in Section \ref{sec:expe}. The conclusion of this work is drawn in Section \ref{conclusion}.

\section{Transform learning}
\label{literature}
\subsection{The transform learning paradigm}

Traditional machine learning methods are limited in their ability to work with raw data. To perform any machine learning task, careful feature engineering is used, which in turn requires domain expertise. Using domain knowledge, a feature extractor is built to transform the raw data into a suitable internal representation. This internal representation is then fed to a learning subsystem, often a classifier to detect a pattern in the input data. Learning the weights between input and representation layer is a challenging task since both weights and output are unknown. This is called representation learning. 

Transform learning, introduced in \cite{Ravishankar2015_part1,Ravishankar2015_part2}, is a representation learning para\-digm that can be viewed as the analysis equivalent of dictionary learning. In dictionary learning, a basis is learned such that it synthesizes the data from the learned coefficients. Transform learning analyzes the data by learning a basis to produce the coefficients. Mathematically this is expressed as $TX \approx Z$, where $T$ is the analysis transform, $X$ is the data, and $Z$ the corresponding coefficient matrix. As proposed in \cite{Ravishankar2013}, the matrices $T$ and $Z$ could be estimated by solving the following optimization problem
\begin{equation}
\operatorname*{minimize}_{T,Z}\; \frac{1}{2} \|TX - Z \|_F^2 + \lambda (\|T\|_F^2 - \log \det T) + \beta \|Z\|_1.
\label{eq:ravi}
\end{equation}
The logarithmic determinant (log det) term aims at imposing a full rank on the learned transform, and at preventing the degenerate solution $T=0, Z=0$. The additional quadratic penalty allows to limit scale indeterminacy. Both these additional penalties improve the conditioning of learnt transforms. Finally, the $\ell_1$ term enforces a sparsity constraint on the coefficients. It is worthy to notice that transform learning is more general than dictionary learning in its notion of compressibility. The learning process is also faster because the sparse coding step reads here as a simple step of thresholding, while in dictionary learning, it requires the resolution of a non trivial optimization problem.

\subsection{Convolutional transform learning}

We proposed in \cite{maggu2018convolutional} the CTL approach, where a set of independent convolution filters are learnt to produce some data representations, in an unsupervised manner. The CTL strategy aims at generating unique and near to orthogonal filters, which in turn produces good features to be used for solving machine learning problems, as we illustrated in our experiments \cite{maggu2018convolutional}. We present here a brief description of this approach, as its notation and concepts will serve as a basis for the deep CTL formulation introduced in this paper. 

We consider a dataset $\left\{x^{(m)}\right\}_{1 \leq m \leq M}$ with $M$ entries in $\mathbb{R}^{N}$. The CTL formulation relies on the key assumption that the representation matrix $T$ gathers a set of $K$ kernels $t_1,\dots,t_K$ with $K$ entries, namely
\begin{equation}
T = [t_1 \;|\;\dots\;|\; t_K]\in\mathbb{R}^{K\times K}.
\end{equation}
This leads to a linear transform applied to the data to produce some features
\begin{equation}
    (\forall m \in \left\{1,\ldots,M\right\}) \qquad  Z_m \approx X^{(m)} T,
\end{equation} 
where $X^{(m)}\in \mathbb{R}^{N \times K}$ are Toeplitz matrices associated to $(x^{(m)})_{1 \leq m \leq M}$ such that
\begin{equation}
    \begin{aligned}
        X^{(m)} T 
        &= \big[ X^{(m)} t_1 \;|\;\dots\;|\; X^{(m)} t_K \big]\\
        &= \big[ t_1  \ast x^{(m)} \;|\dots|\; t_K  \ast x^{(m)} \big],
    \end{aligned}
\end{equation}
and $\ast$ is a discrete convolution operator with suitable padding.
Let us denote
\begin{equation}
Z = \begin{bmatrix}
Z_1\\
\vdots\\ 
Z_M
\end{bmatrix}\in\mathbb{R}^{NM\times K}.
\end{equation}
The goal is then to estimate $(T,Z)$ from $\left\{x^{(m)}\right\}_{1 \leq m \leq M}$. To do so, we proposed in \cite{maggu2018convolutional} a penalized formulation of the problem, introducing suitable conditioning constraints on the transforms, and sparsity constraint on the coefficients. The learning of $(T,Z)$ was then performed using an alternating minimization scheme with sounded convergence guarantees. The aim of the present paper is to introduce a multi-layer formulation of the CTL, in order to learn deeper representations, with the aim of improving the representation power of the features. 

\section{Proposed Approach}
\label{proposed}
\subsection{Deep convolutional transform model}
Starting from the CTL model, we propose to stack several layers of it to obtain a deep architecture. For every $\ell \in \left\{1,\ldots,L\right\}$, we will seek for the transform matrix
\begin{equation}
    T_\ell = \left[t_{1,\ell} \;|\dots|\; t_{K,\ell}\right] \in\mathbb{R}^{K\times K},
\end{equation}
where $t_{k,\ell} \in\mathbb{R}^{K}$ is the $k$-th kernel on the $\ell$-th layer of the representation. The associated coefficients will be denoted as
\begin{equation}
Z_\ell =
\begin{bmatrix}
Z_{1,\ell }\\
\vdots\\ 
Z_{M,\ell}
\end{bmatrix}
\in \mathbb{R}^{NM \times K},
\end{equation}
with
\begin{equation}
(\forall m \in \left\{1,\ldots,M\right\})\qquad Z_{m,\ell} = \big[z_1^{(m,\ell)} \;|\dots|\; z_K^{(m,\ell)} \big] \in \mathbb{R} ^{N \times K}.
\end{equation}
The learning of $(T_\ell)_{1 \leq \ell \leq L}$ and $(Z_\ell)_{1 \leq \ell \leq L}$ will be performed by solving
\begin{equation}
\operatorname*{minimize}_{(T_\ell)_{1 \leq \ell \leq L}, (Z_\ell)_{1 \leq \ell \leq L}}\; F(T_1,\ldots,T_L, Z_1,\ldots,Z_L)
\label{eq:multi_layer}
\end{equation}
where
\begin{multline}
    F(T_1,\ldots,T_L, Z_1,\ldots,Z_L) = 
    \sum\limits_{\ell  = 1}^L \left( \frac{1}{2}\sum\limits_{m = 1}^M {||{{\cal Z}_{m,\ell  - 1}}{T_\ell } - {Z_{m,\ell }}||_F^2}  + \mu ||{T_\ell }||_F^2 \right.\\
    \left. \phantom{\sum\limits_{m = 1}^M} 
    - \lambda \log \det ({T_\ell }) + \beta ||{Z_\ell }|{|_1} + {\iota _{+}}({Z_\ell }) \right),
\end{multline}
Here, we denote $\iota_+$ the indicator function of the positive orthant, equals to 0 if all entries of its input have non negative elements, and $+ \infty$ otherwise. Moreover, by a slight abuse of notation, we denote as $\log \det$ the sum of logarithms of the singular values of a squared matrix, taking infinity value as soon as one of those is non positive.
The first layer follows the CTL strategy, that is $\mathcal{Z}_{m,0} \equiv X^{(m)}$. Moreover, for every $\ell \in \left\{2,\ldots,L\right\}$, we introduced the linear operator $\mathcal{Z}_{m,\ell-1}$ so as to obtain the compact notation for the multi-channel convolution product:
\begin{align}
    \mathcal{Z}_{m,\ell-1} T_{\ell}  & = \left[\mathcal{Z}_{m,\ell-1} t_{1,\ell} | \, \ldots \, | \mathcal{Z}_{m,\ell-1} t_{K,\ell} \right] \\
    & = \left[t_{1,\ell} \ast z_1^{(m,\ell-1)} | \ldots | t_{K,\ell} \ast z_K^{(m,\ell-1)} \right].
\end{align}
\subsection{Minimization algorithm}
Problem \eqref{eq:multi_layer} is non-convex. However it presents a particular multi-convex structure, that allows us to make use of an alternating proximal minimization algorithm to solve it~\cite{Attouch_Bolte_2011,Chouzenoux15jogo}. The proximity operator \cite{Combettes_Book_10} of a proper, lower semi-continuous, convex function $\psi: \mathcal{H} \mapsto ]-\infty,+\infty]$, with $(\mathcal{H},\|\cdot\|)$ a normed Hilbert space, is defined as\footnote{See also http://proximity-operator.net/}
\begin{equation}\label{eq_pbprox}
(\forall \widetilde{X} \in \mathcal{H})\quad \operatorname{prox}_{\psi}(\widetilde{X}) = \operatorname*{argmin}_{X\in \mathcal{H}}\; \psi(X) + \frac{1}{2}\|X - \widetilde{X} \|^2. 
\end{equation}
The alternating proximal minimization algorithm then consists in performing iteratively proximity updates, on the transform matrix, and on the coefficients. The iterates are guaranteed to ensure the monotonical decrease of the loss function $F$. Convergence to a local minimizer of $F$ can also be ensured, under mild technical assumptions. The algorithm reads as follows:
\begin{equation}
\begin{array}{*{20}{l}}
{{\rm{For}}\quad i = 0,1, \ldots }\\
\left\lfloor {\begin{array}{*{20}{l}}
{{\rm{For}}\quad \ell  = 1, \ldots ,L}\\
\left\lfloor {\begin{array}{*{20}{l}}
{T_\ell ^{[i + 1]}}&{ = {{{\mathop{\rm prox}\nolimits} }_{{\gamma _1}F(T_1^{[i + 1]}, \ldots ,T_L^{[i]},Z_1^{[i + 1]}, \ldots ,Z_L^{[i]})}}\left( {T_\ell ^{[i]}} \right)}\\
{Z_\ell ^{[i + 1]}}&{ = {{{\mathop{\rm prox}\nolimits} }_{{\gamma _2}F(T_1^{[i + 1]}, \ldots ,T_L^{[i]},Z_1^{[i + 1]}, \ldots ,Z_L^{[i]})}}\left( {Z_\ell ^{[i]}} \right)}
\end{array}}\right.
\end{array}}\right.
\end{array}
\end{equation}
with $T_\ell^{[0]} \in \mathbb{R}^{K\times K}$, $Z_\ell^{[0]} \in \mathbb{R}^{NM\times K}$, and $\gamma_1$ and $\gamma_2$ some positive constants. We provide hereafter the expression of the proximity operators involved in the algorithm, whose proof are provided in the appendix.

\subsubsection{Update of the transform matrix:}

Let $i \in \mathbb{N}$ and $\ell \in \left\{1,\ldots,L\right\}$. Then
\begin{equation}
\begin{aligned}
{T_\ell ^{[i + 1]}}& = {{{\mathop{\rm prox}\nolimits} }_{{\gamma _1}F(T_1^{[i + 1]}, \ldots ,T_L^{[i]},Z_1^{[i + 1]}, \ldots ,Z_L^{[i]})}}\left( {T_\ell ^{[i]}} \right), \\
&= \operatorname*{argmin}_{{T_\ell } \in {\mathbb{R} ^{K \times K}}}\;
\frac{1}{{2{\gamma _1}}}||{T_\ell } - T_\ell ^{[i]}||_F^2 \\
&\qquad\qquad\quad  +\frac{1}{2}\sum\limits_{m = 1}^M {||{\cal Z}_{m,\ell  - 1}^{[i + 1]}{T_\ell } - Z_{m,\ell }^{[i]}||_F^2}  + \mu ||{T_\ell }||_F^2 - \lambda \log \det ({T_\ell })
\\
& = \frac{1}{2}{\Lambda ^{ - 1}}V\left( {\Sigma  + {{({\Sigma ^2} + 2\lambda \text{Id} )}^{1/2}}} \right){U^ \top },
\end{aligned}   
\end{equation}
with
\begin{equation}
    \Lambda^\top \Lambda = \sum_{m=1}^M (\mathcal{Z}_{m,\ell-1}^{[i+1]})^\top (\mathcal{Z}_{m,\ell-1}^{[i+1]}) + (\gamma_1^{-1} + 2 \mu) \text{Id}.
\end{equation}
Hereabove, we considered the singular value decomposition:
\begin{equation}
U \Sigma V^\top  =  \left(\sum_{m=1}^M (Z_{m,\ell}^{[i]})^\top (\mathcal{Z}_{m,\ell-1}^{[i+1]}) + \gamma_1^{-1} T_{\ell}^{[i]}\right)\Lambda^{-1}.
\end{equation}

\subsubsection{Update of the coefficient matrix:}

Let $i \in \mathbb{N}$. We first consider the case when $\ell \in \left\{1, \ldots,L-1\right\}$ (recall that $\mathcal{Z}_{m,0} = X^{(m)}$ when $\ell = 1$). Then
\begin{equation}
\!\!\!\!\!\!\!
\begin{aligned}
{Z_\ell ^{[i + 1]}}&{ = {{{\mathop{\rm prox}\nolimits} }_{{\gamma _2}F(T_1^{[i + 1]}, \ldots ,T_L^{[i]},Z_1^{[i + 1]}, \ldots ,Z_L^{[i]})}}\left( {Z_\ell ^{[i]}} \right)},\\
&= 
\operatorname*{argmin}_{{Z_\ell } \in {\mathbb{R} ^{MN \times K}}}\;
\frac{1}{{2{\gamma _2}}}||{Z_\ell } - Z_\ell ^{[i]}||_F^2 \\
&\qquad\qquad\quad +\frac{1}{2}\sum\limits_{m = 1}^M {||{\cal Z}_{m,\ell  - 1}^{[i + 1]}T_\ell ^{[i + 1]} - {Z_{m,\ell }}||_F^2} \\
& \qquad\qquad\quad + \frac{1}{2}\sum\limits_{m = 1}^M {||{{\cal Z}_{m,\ell }}T_{\ell  + 1}^{[i + 1]} - Z_{m,\ell  + 1}^{[i]}||_F^2} 
\\
& \qquad\qquad\quad + \beta ||{Z_\ell }|{|_1} + {\iota _+}({Z_\ell }).
\end{aligned}
\end{equation}
Although the above minimization does not have a closed-form expression, it can be efficiently carried out with the projected Newton method.
In the case when $\ell = L$, the second term is dropped, yielding
\begin{equation}
\begin{aligned}
Z_L^{[i+1]} &= \operatorname{prox}_{\gamma_2 F(T_1^{[i+1]},\ldots,T_{L}^{[i+1]},Z_{1}^{[i+1]},\ldots,Z_{L-1}^{[i+1]},\cdot)} \left(Z_L^{[i]}\right) \\
&= \operatorname*{argmin}_{Z_L \in \mathbb{R}^{MN\times K}}\;
\frac{1}{2 \gamma_2} \|Z_L - Z_L^{[i]} \|_F^2 \\
& \qquad\qquad\quad + \frac{1}{2} \sum_{m=1}^M \|\mathcal{Z}_{m,L-1}^{[i+1]} T_{L}^{[i+1]}  - Z_{m,L}\|_F^2 + \beta \|Z_L\|_1 + \iota_{+}(Z_L).
\end{aligned}
\end{equation}
Hereagain, the projected Newton method can be employed for the minimization.

\section{Numerical results}
\label{sec:expe}

\subsection{Datasets}
To assess the performance of the proposed approach, we considered the following image datasets\footnote{http://www.cad.zju.edu.cn/home/dengcai/Data/FaceData.html} of small-to-medium size.
 

\begin{enumerate}
\item \emph{YALE \cite{belhumeur1997yale}:} The Yale dataset contains 165 images of 15 individuals, downscaled to 32-by-32 pixels. There are 11 images per subject, one per different facial expression or configuration. For our experiments, we shuffled all the samples, and took 70\% for training and 30\% for testing. Moreover, we full-size YALE images of size 150-by-150 pixels.

\item \emph{E-YALE-B \cite{lee2005acquiring}:} The extended Yale B database contains 2432 images with 38 subjects under 64 illumination conditions. Each image is cropped to 192-by-168 pixels and downscaled to 48-by-42 pixels. For our experiments, we shuffled all the samples, took 70\% for training and 30\% for testing.

\item \emph{AR-Face \cite{martinez1998ar}:} This database contains more than 4000 images of 126 different subjects (70 male and 56 female). The images have various facial expressions, the lighting varies, and some of the images are partially occluded by sunglasses and scarves. For our experiments, we selected 2600 images of 100 individuals (50 males and 50 females), that is 26 different images for each subject. Train set contains 2000 images, and 600 images are kept in the test set. Each image has 540 features.
\end{enumerate}

\subsection{Numerical results}

\begin{table}[htbp]
\caption{Accuracy on SVM with layers}
\begin{center}
    \begin{tabular}{|c|c|c|c|c|}
    \hline
         \textbf{Dataset} & \textbf{CTL} & \textbf{DCTL-2} & \textbf{DCTL-3} & \textbf{DCTL-4} \\ \hline
         \textbf{YALE 150 $\times$ 150} & 94.00 & 94.28 & \textbf{96.00} & 92.21 \\
         \textbf{YALE 32 $\times$ 32} & 88.00 & 89.11 & \textbf{90.00} & 87.73 \\
         \textbf{E-YALE-B} & 97.38 & 97.00 & \textbf{98.00} & 94.44 \\
         \textbf{AR-Faces} & 88.87 & 92.22 &  \textbf{97.67} & 82.21 \\ \hline
    \end{tabular}
    \label{tab:deep}
\end{center}
\end{table}
\begin{table}[htbp]
\caption{Classification Accuracy using KNN}
\begin{center}
\begin{tabular}{|c|c|c|c|}
\hline
     \textbf{Dataset} & \textbf{Raw Features} & \textbf{CTL} & \textbf{DCTL} \\ \hline
     \textbf{YALE 150 $\times$ 150} & 78.00 &  70.00 & \textbf{80.00} \\
     \textbf{YALE 32 $\times$ 32} & \textbf{60.00}  & 58.85 & \textbf{60.00} \\
     \textbf{E-YALE-B} & 71.03  & 84.00 & \textbf{85.00} \\
     \textbf{AR-Faces} & 55.00 & \textbf{56.00} & 58.00 \\
     \hline
\end{tabular}
\label{tab:knn_classification}
\end{center}
\end{table}
\begin{table}[htbp]
\caption{Classification Accuracy using SVM}
\begin{center}
    \begin{tabular}{|c|c|c|c|}
    \hline
         \textbf{Dataset} & \textbf{Raw Features} & \textbf{CTL} & \textbf{DCTL} \\ \hline
         \textbf{YALE 150 $\times$ 150} & 93.00 &  94.00 & \textbf{96.00} \\
         \textbf{YALE 32 $\times$ 32} & 68.00  & 88.00 & \textbf{90.00} \\
         \textbf{E-YALE-B} & 93.24  & 97.38 & \textbf{98.00}\\
         \textbf{AR-Faces} & 87.33 & 88.87 & \textbf{97.67} \\ \hline
    \end{tabular}

    \label{tab:svm_classification}
\end{center}
\end{table}
\begin{table}[htbp]
\caption{Convolutional Transformed Clustering: ARI}
\begin{center}
\begin{tabular}{|c|c|c|c|}
\hline
     \textbf{YALEB/Method} & \textbf{Raw Features} & \textbf{DCTL-2} & \textbf{DCTL-3} \\ \hline
     \textbf{K-means} & 0.785 & 0.734 & \textbf{0.788} \\
     \textbf{Random} & 0.733 & 0.718 & \textbf{0.738} \\
     \textbf{PCA-based} & 0.734 & \textbf{0.791} & 0.777\\ \hline
    \end{tabular}
\label{tab:clustering}
\end{center}
\end{table}
\begin{table}[htbp]
\caption{Clustering time in sec}
\begin{center}
\begin{tabular}{|c|c|c|c|}
\hline
     \textbf{YALEB/Method} & \textbf{Raw Features} & \textbf{DCTL-2} & \textbf{DCTL-3} \\ \hline
     \textbf{K-means} & 2.28 & 0.45 & \textbf{0.14} \\
     \textbf{Random} & 1.95 & 0.33 & \textbf{0.08} \\
     \textbf{PCA-based }& 0.36 & 0.09 & \textbf{0.03}\\ \hline
\end{tabular}
\label{tab:clustering time}
\end{center}
\end{table}

We experiment on the YALE, EYALEB and AR faces datasets; these are well known benchmarking face datasets. 
In the first set of experiments, we want to show that the accuracy of deep transform learning indeed improves when one goes deeper. Going deep beyond three layers makes performance degrade as the model tends to overfit for the small training set. To elucidate, we have used a simple support vector machine (SVM) classifier. The results are shown in Table \ref{tab:deep} for levels 1, 2, 3 and 4. It has already been shown in \cite{maggu2018convolutional} that the single layer CTL yielded better results than other single layer representation learning tools, including dictionary learning and transform Learning. Therefore it is expected that by going deeper, we will improve upon their deeper counterparts. We do not repeat those baseline experiments here, by lack of space. We also skip comparison with CNNs because of its supervised nature, whereas the proposed technique is unsupervised. We only show comparison of our proposed technique with raw features and with CTL. We take extracted features from the proposed DCTL and perform classification using two classifiers, namely KNN and SVM. The classification accuracy is reported in table \ref{tab:knn_classification} and table \ref{tab:svm_classification}. Then we perform clustering on the extracted features of DCTL and report the comparison of Adjusted Rank Index (ARI) in table \ref{tab:clustering}. We also report clustering time on extracted features in table \ref{tab:clustering time}. It is worthy to remark that the time to cluster extracted features from the proposed methodology is comparatively less than others. 

\section{Conclusion}
\label{conclusion}
This paper introduces a deep representation learning technique, named Deep Convolutional Transform Learning. Numerical comparisons are performed with the shallow convolutional transform learning formulations on image classification and clustering tasks. In the future, we plan to compare with several other deep representation learning techniques, namely stacked autoencoder and its convolutional version, restricted Boltzmann machine and its convolutional version, discriminative variants of deep dictionary and transform Learning.

\section{Appendix: Proofs of the proximity updates}

\subsection{Update of $T$}

Let us consider $M=1$ for simplicity, but note that all the calculations hold for $M$ greater than $1$. We want to minimize a function of the form:
\begin{equation}
\Phi(T) = \frac{1}{2} ||X T - Z||_F^2 
    + \mu ||T||_F^2 - \lambda \log \det (T) + \frac{1}{{2{\gamma _1}}}||T - {T^{[n]}}||_F^2.
\end{equation}

Using some linear algebra, We can easily prove that:
\begin{equation}
\Phi(T) = \frac{1}{2}||{W^{1/2}}T - Y||_F^2 - \lambda \log \det (T) + c
\end{equation}
with $c$ a constant with respect to $T$,
\begin{equation}
    W = {X^\top}X + (2\mu + \frac{1}{\gamma _1}) {\text{Id}}
\end{equation}
and
\begin{equation}
    Y = {W^{ - 1/2}}({Z^\top}X + \frac{1}{\gamma _1}{T^{[n]}}).
\end{equation}
Since $W$ is invertible, one can perform the change of variable $\tilde T = {W^{1/2}}T$, that is $T = {W^{ - 1/2}}\tilde T$. Thus,
\begin{equation}
    \operatorname{argmin}_T \Phi(T) = {W^{ - 1/2}}\operatorname{argmin}_T {\tilde{T}} \Phi({W^{ - 1/2}}\tilde T),
\end{equation}
with
\begin{equation}
    \Phi({W^{ - 1/2}}\tilde T) = \frac{1}{2}||\tilde T - Y||_F^2 - \lambda \log \det ({W^{ - 1/2}}\tilde T) + c.
\end{equation}
Moreover,
\begin{equation}
    \log \det ({W^{ - 1/2}}\tilde T) = \log \det (\tilde T).
\end{equation}
Thus, 
\begin{equation}
    \operatorname{argmin}_{\tilde T} \Phi({W^{ - 1/2}}\tilde T) = \operatorname{prox}_{\lambda \log \det(\tilde T)}(Y),
\end{equation}
which maps with the proximity operator of the logarithmic determinant function with weight $\lambda$. We can then apply \cite[Example 24.66]{Combettes_Book_10} and \cite[Proposition 24.68]{Combettes_Book_10} to conclude the proof.

\subsection{Update of $Z$}
We have to solve:
\begin{multline}
    \operatorname{argmin}_{Z} \frac{1}{2}\sum\limits_{m = 1}^M \|{X^{(m)}}{T^{[n + 1]}} - {Z_m}\|_F^2 
    + \beta ||Z|{|_1} + {\iota _+}(Z) + \frac{1}{{2{\gamma _2}}}||Z - {Z^{[n]}}||_F^2.
    \label{Z_i+1}
\end{multline}
 The function in \eqref{Z_i+1} is fully separable, i.e. it can be written as a sum over all the entries of matrix $Z$. Due to the separability property of the proximity operator, it is sufficient to resonate on the minimization of scalar function with respect to $Z_{p,q,r}$: 
\begin{equation}
\frac{1}{2}{({[{X^m}{T^{i + 1}}]_{p,q}} - {Z_{p,q,r}})^2} + \beta |{Z_{p,q,r}}| + {\iota_+}({Z_{p,q,r}}) + \frac{1}{2}{\gamma _2}{({Z_{p,q,r}} - Z_{p,q,r}^i)^2}.
\end{equation}
One can conclude, noticing that the term $\beta |\cdot| + \iota_+$ corresponds to case 'ix' of  in  \cite[Table 10.2]{Combettes_Book_10}, and by applying case 'iv' of \cite[table 10.1]{Combettes_Book_10} to process the final quadratic term.

%
%
%
\bibliographystyle{splncs04}
\bibliography{IEEEexample}
%




\end{document}